\documentclass[conference]{IEEEtran}
\IEEEoverridecommandlockouts
\usepackage{cite}
\usepackage{amsmath,amssymb,amsfonts}
\usepackage{algorithmic}
\usepackage{graphicx}
\usepackage{subcaption}
\usepackage[font=small]{caption}
\usepackage{textcomp}
\usepackage{xcolor}

\usepackage{listings}
\usepackage{tikz}
\usepackage{textcomp}
\usepackage{hyperref}
\usepackage{lipsum}

\newcommand\copyrighttext{%
  \footnotesize \textcopyright 2025 IEEE. Personal use of this material is permitted.
  Permission from IEEE must be obtained for all other uses, in any current or future
  media, including reprinting/republishing this material for advertising or promotional
  purposes, creating new collective works, for resale or redistribution to servers or
  lists, or reuse of any copyrighted component of this work in other works.
  DOI: \href{<https://ieeexplore.ieee.org/abstract/document/11114174>}{10.1109/CoG64752.2025.11114174}}
\newcommand\copyrightnotice{%
\begin{tikzpicture}[remember picture,overlay]
\node[anchor=south,yshift=10pt] at (current page.south) {\fbox{\parbox{\dimexpr\textwidth-\fboxsep-\fboxrule\relax}{\copyrighttext}}};
\end{tikzpicture}%
}

\def\BibTeX{{\rm B\kern-.05em{\sc i\kern-.025em b}\kern-.08em
    T\kern-.1667em\lower.7ex\hbox{E}\kern-.125emX}}
\begin{document}
\bstctlcite{IEEEexample:BSTcontrol}

\newcommand\blfootnote[1]{%
  \begingroup
  \renewcommand\thefootnote{}%
  \footnote{#1}%
  \addtocounter{footnote}{-1}%
  \endgroup
}

\newcommand{\vspacereducer}{-0.3cm}

\title{Learning Representations in Video Game Agents with Supervised Contrastive Imitation Learning \\

}

\author{\IEEEauthorblockN{ Carlos Celemin, Joseph Brennan, Pierluigi Vito Amadori, Tim Bradley}
\textit{Sony Interactive Entertainment 
Europe}\\
London, United Kingdom \\
\{carlos.celemin, joe.brennan, pierluigi.vito.amadori, timothy.bradley\}@sony.com

}

\maketitle
\copyrightnotice

\begin{abstract}
This paper introduces a novel application of Supervised Contrastive Learning (SupCon) to Imitation Learning (IL), with a focus on learning more effective state representations for agents in video game environments. 
The goal is to obtain latent representations of the observations that capture better the action-relevant factors, thereby modeling better the cause-effect relationship from the observations that are mapped to the actions performed by the demonstrator, for example, “the player jumps whenever an obstacle appears ahead.”
We propose an approach to integrate the SupCon loss with continuous output spaces, enabling SupCon to operate without constraints regarding the type of actions of the environment. 
Experiments on the 3D games Astro Bot and Returnal, and multiple 2D Atari games show improved representation quality, faster learning convergence, and better generalization compared to baseline models trained only with supervised action prediction loss functions.

\end{abstract}


\section{Introduction}
Automated game testing has increasingly employed the use of Machine Learning (ML) agents\cite{zheng2019wuji,bergdahl2020augmenting,tufano2022using,amadori2024robust,celemin2024bayesian}. 
In more general research contexts, Imitation Learning (IL) has been used to train agents to play video games\cite{thurau2004imitation,harmer2018imitation,yadgaroff2024improving}. 
Training agents becomes more difficult when they are limited to high-dimension visual inputs, rather than structured, low-dimensional internal game states, typically used in prior work.

For unreleased titles, game developers can provide ML agents with access to detailed game state data instead of raw pixel observations. 
However, when deploying agents to test published games, backward compatibility becomes a significant issue: the game code is not actively maintained, and internal state access may be restricted or unavailable. 
In such cases, training agents in an end-to-end manner from visual input becomes the only viable approach.

These constraints often require larger datasets and longer training durations, increasing the risk of overfitting. 
This makes it crucial to focus on learning effective state representations, that identify the key input patterns associated with demonstrated actions and that generalize well to previously unseen game states.

\begin{figure}
    \centering
    \includegraphics[trim={1.5cm 7cm 15cm 8.5cm},clip, width=0.955\linewidth]{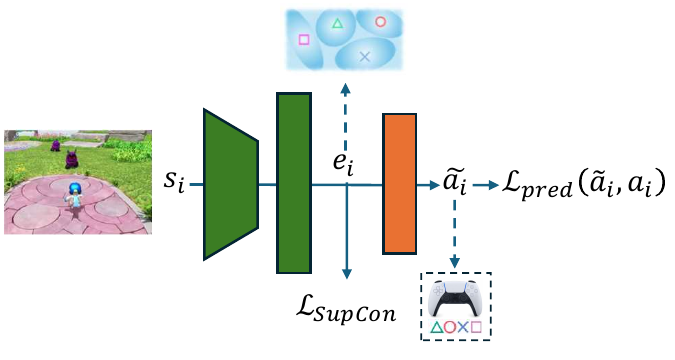}
    \caption{Architecture for training a policy with Supervised Contrastive Imitation Learning. }
    \label{fig:diagram}
    
\vspace{\vspacereducer}
\end{figure}

This work focuses on learning state representations for Imitation Learning (IL) agents that play video games. 
Our goal is to train feature extractors that embed observations into representations that facilitate decision-making (Fig. \ref{fig:diagram}).
To this end, we explore how to introduce inductive biases that constrain the solution space toward more meaningful representations. 
In recent years, Self-Supervised Learning (SSL) methods \cite{jaiswal2020survey,chen2020simple} have become widely used for learning general-purpose representations.
These methods typically leverage geometric data augmentations to form positive and negative pairs: similar inputs are pulled together in embedding
space, while dissimilar ones are pushed apart. However, in video game environments, precise spatial configurations are crucial, not just object identity. 
The arrangement of game entities directly impacts the agent’s behavior, such as aiming at a target, dodging bullets, or timing a jump over an obstacle at the right distance.

Various studies in State Representation Learning (SRL) have proposed auxiliary losses involving both states and actions, including forward models, inverse dynamics models, and objectives that relate temporal changes in state to those in actions \cite{lesort2018state}. 
These approaches introduce useful inductive biases and have shown improvements in generalization for decision-making agents. 
However, they often fall short of the core goal: learning representations explicitly shaped by the agent’s actions, rather than simply influenced by temporal or physical dynamics.

\begin{figure}[htbp]
  \centering
  \begin{subfigure}{0.48\columnwidth}
    \centering
    \includegraphics[width=\linewidth]{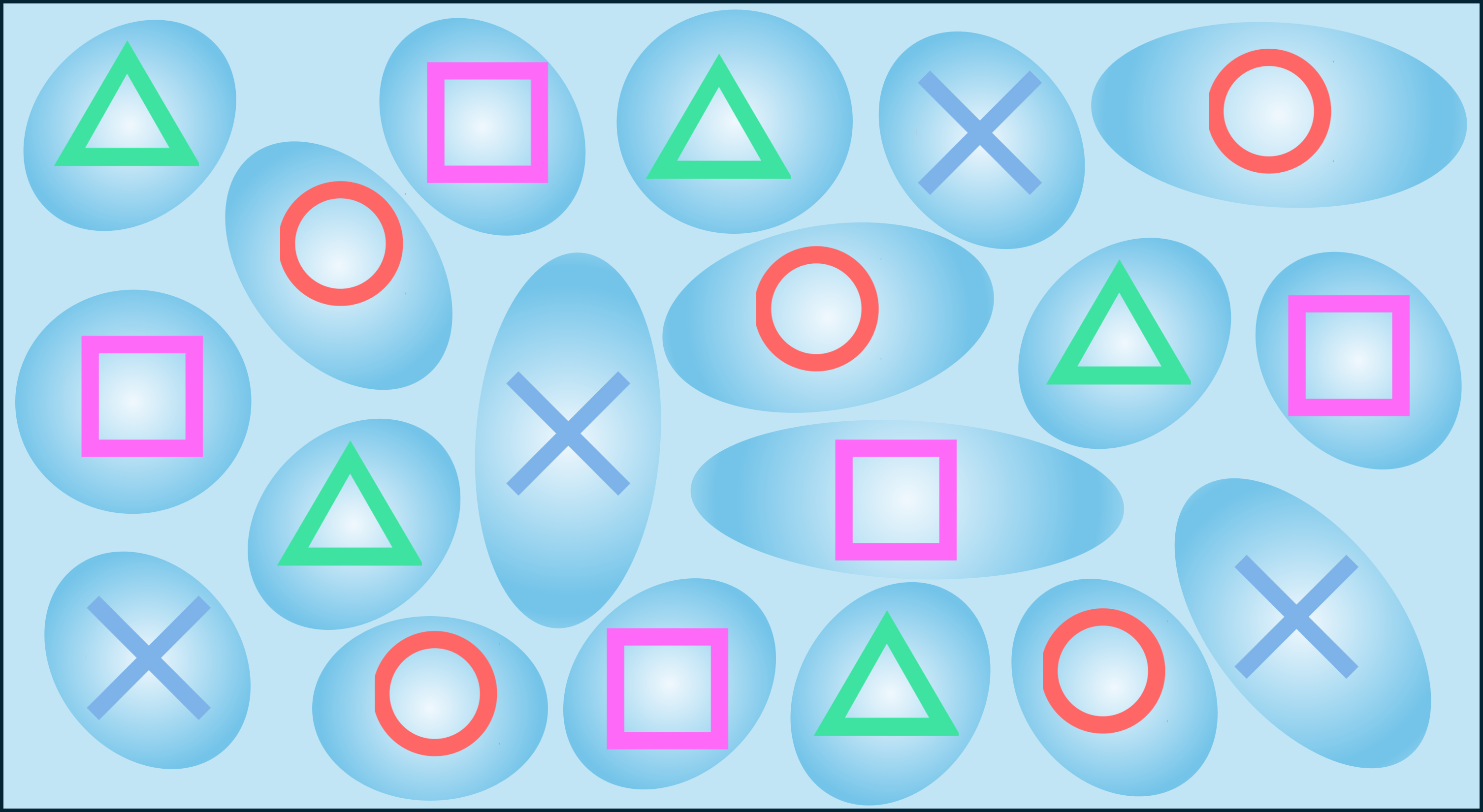}
    \caption{ Only with the supervised learning losses.}
    \label{fig:fig1a}
  \end{subfigure}
  \hfill
  \begin{subfigure}{0.48\columnwidth}
    \centering
    \includegraphics[width=\linewidth]{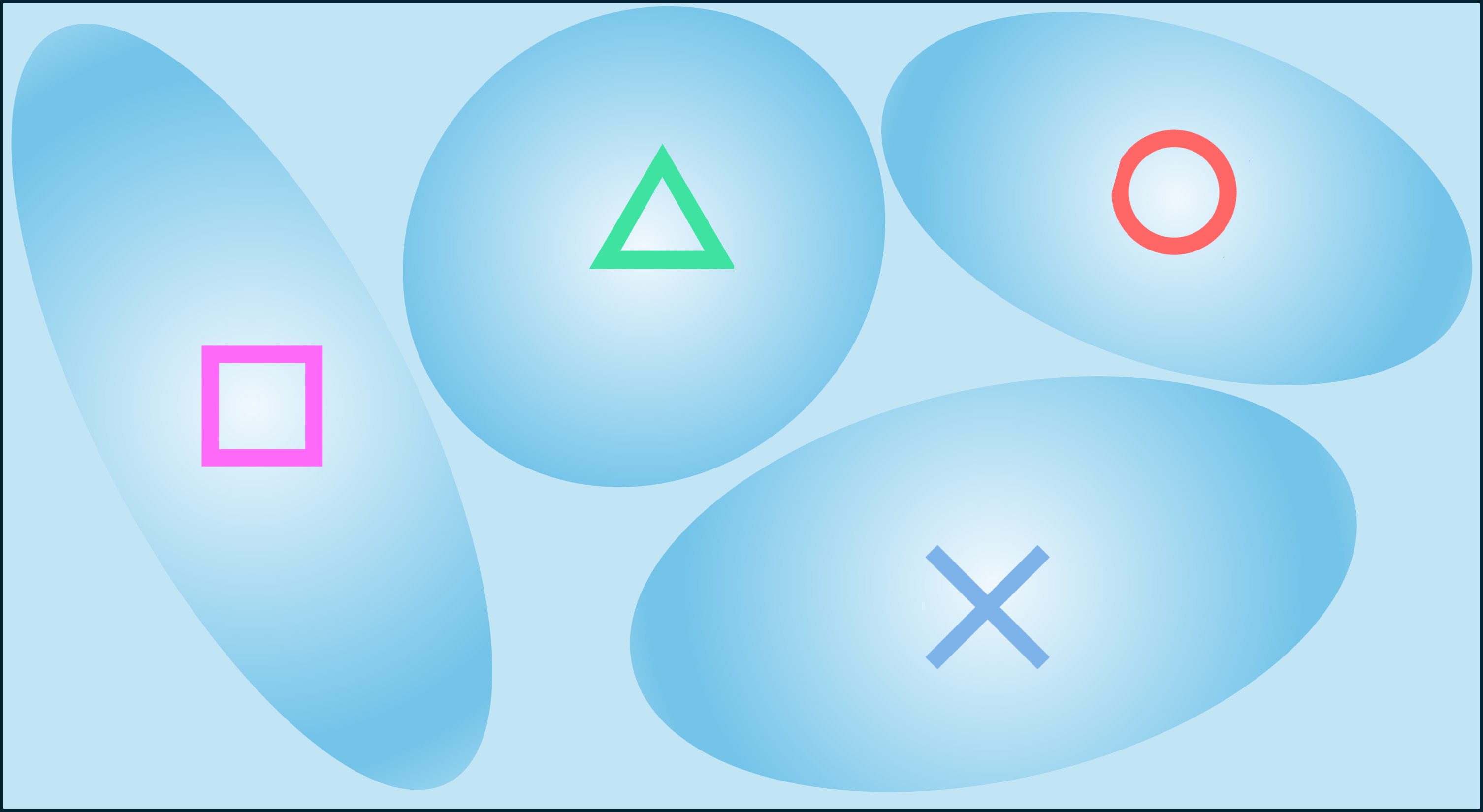}
    \caption{Structured with action-relevant factors.}
    \label{fig:fig1b}
  \end{subfigure}
  \caption{Learned embedding spaces.}
  \label{fig:embeddings}
\vspace{\vspacereducer}
\end{figure}

Motivated by this, we aim to learn state representations that reside in a latent space where actions are easily distinguishable. 
In other words, the geometry of the latent space should be structured according to the actions associated with each observation (as in Fig. \ref{fig:embeddings}). 
The underlying assumption is that observations leading to the same action share common factors of variation and should therefore be embedded closely together, while embeddings of observations tied to different actions should be well-separated. 
If the representation can successfully capture these action-relevant factors, IL agents can be trained more efficiently and with better generalization. 

To this end, we propose incorporating a Supervised Contrastive (SupCon) loss \cite{khosla2020supervised} into the IL training pipeline as a regularizer, called Supervised Contrastive Imitation Learning (SCIL). 
Similar in spirit to contrastive SSL approaches, but without relying on augmentations which can distort spatial information critical to decision-making, this method uses action labels to determine positive and negative sample pairs. 
This conditions the latent space directly on the task-relevant output: the action. 
Although SupCon has been widely used in supervised learning settings, to our knowledge, this is the first application of SupCon in Imitation Learning.

Experimental results demonstrate that adding SupCon to the training process improves convergence, allowing agents to reach higher performance metrics earlier in training.
This proposed approach is architecture-agnostic and can be easily adopted within other types of IL strategies.

\section{Method}\label{sec:method}
In this work, we combine the supervised losses directly computed with the output of the policy network, and the SupCon loss in equation \eqref{eq:SupCon} computed in the embedding space after the feature extractor layers. 
In Fig. \ref{fig:diagram}, the diagram of the system is depicted, in which $L_{pred}$ is the predictive task loss used for supervised learning. 
It could be used Cross Entropy, MSE, or a contrastive loss like the InfoNCE used for training the Energy Based Model (EBM) in \cite{amadori2024robust}, which is one of the  baseline models used in our experiments.

\begin{equation}\label{eq:SupCon}
\mathcal{L} = \sum_{i \in I} \frac{-1}{|P(i)|} \sum_{p \in P(i)} \log \frac{\exp(\mathbf{e}_i \cdot \mathbf{e}_p / \tau)}{\sum_{a \in A(i)} \exp(\mathbf{e}_i \cdot \mathbf{e}_a / \tau)}
\end{equation}
where:
\begin{itemize}
  \item $I$ is the set of all indices in the batch.
  \item $P(i)$ is the set of indices of positives for anchor $i$ (i.e., same class as $i$, excluding $i$).
  \item $A(i)$ is the set of all indices in the batch except $i$.
  \item $\mathbf{e}_i$ is the normalized embedding of sample $i$.
  \item $\tau$ is the temperature scaling parameter.
\end{itemize}

Since the embeddings $\mathbf{e}_i$ in \eqref{eq:SupCon} are normalized, the products there are the implementation of the cosine similarity.

Several considerations must be addressed to adapt SupCon for IL in the context of video games, as the original SupCon formulation does not directly align with these settings. Notably, the action spaces in video games are often multidimensional, comprising a mixture of continuous and discrete variables. This complexity requires modifications to how positive and negative pairs are defined and selected during training, as similarity between actions must account for partial overlaps or close values in continuous dimensions, rather than strict class equality.

\subsection{No artificial views or positive pair augmentations}
The original SupCon implementation assumes that each sample is accompanied by multiple augmented views to form positive pairs, a common practice in contrastive learning, particularly in benchmarks using standard classification datasets.

However, such augmentations are not well-suited for sequential decision-making tasks like imitation learning in video games. In these environments, spatial information is as important as visual patterns or object shapes. Applying geometric augmentations such as shifting, mirroring, rotation, or scaling can distort critical spatial cues, leading to degraded performance of the learned policy. These transformations may violate semantic consistency between states and actions, eventually introducing noise into the representation learning process.

Instead, in our approach, action (label) information alone is used to determine positive sample pairs. This will be elaborated on in Sec. \ref{sec:label_mapping}

\subsection{Generation of Categorical Labels from Continuous Actions}
During training, the models developed in this work can be interpreted as a combination of a multi-label classifier (for discrete action heads) and a regression model (for continuous actions). The Supervised Contrastive Learning framework is originally designed for discrete classification tasks, where it assumes that samples belong to the same class or not, i.e., the notion of similarity is binary. However, this assumption does not naturally extend to settings involving continuous outputs, where actions can be similar to varying degrees rather than being strictly equivalent or different.

This raises a key question: How can categorical labels be generated in a setting where part of the action space is continuous? Addressing this requires strategies that define positive pairs not only by class equality but also by similarity thresholds or quantization schemes that respect the structure of the continuous action space.

\subsubsection{Continuous Space Discretization}

We propose discretizing each continuous action dimension into $B$ bins. This allows us to define similarity in the continuous space by treating nearby values within the same bin as equivalent, effectively assigning them to the same "class." As a result, we obtain a proxy output (or action) vector composed entirely of categorical variables across all action dimensions.
This discretization is applied only for the purpose of computing the SupCon loss, the continuous action prediction at the model’s output remains treated as a continuous regression problem.

\subsubsection{Mapping from Multi-Label to Single Categorical Label}\label{sec:label_mapping}

Once the continuous action dimensions have been discretized, the resulting proxy action vector aligns with a multi-label classification format. To simplify the contrastive learning setup, we group samples that share the same proxy action vector into a single class. The assumption here is that samples with identical discretized action vectors represent functionally similar behaviors and should be treated as positive pairs.

To convert multi-dimensional discrete vectors into single categorical labels, we implement a positional encoding scheme inspired by numeric systems (e.g., binary or decimal). However, unlike uniform-base systems, each position (i.e., action dimension) in our case may have a different base, determined by the number of bins used in that dimension. This system allows us to map each unique combination of discretized values to a distinct class index, as expected by the SupCon loss formulation.

Let $\mathbf{v} = (v_1, v_2, \ldots, v_D)$ be the vector of discretized action values, where each $v_d \in \{0, 1, \ldots, B_d - 1\}$, and $B_d$ is the number of bins in the $d$-th dimension. We use a mixed-radix positional encoding that maps $\mathbf{v}$ to a unique integer label $L$ as:
\begin{equation}
    L = \sum_{d=1}^{D} v_d \cdot \left( \prod_{k=1}^{d-1} B_k \right); \quad \prod_{k=1}^{0} B_k = 1.
\end{equation}\label{eq:PE}
\vspace{\vspacereducer}

\subsection{Handling the Absence of Positive Pairs in a Mini-Batch}
The original implementation of the SupCon loss does not account for the edge case in which a sample lacks a positive pair within a mini-batch, leading to a division by zero. 
This issue does not arise in the original SupCon setting, where each sample is augmented to create at least one positive pair. 
As a result, regardless of the batch composition, every anchor always has at least one positive.

In contrast, our domain, where data augmentations are intentionally avoided, allows for the possibility that some samples may not have any positive counterparts in the mini-batch. 
To address this, we introduced additional safeguards in our implementation (which is based on the original implementation \cite{khosla2020supervised}) to handle such cases explicitly, avoiding numerical instability.
That said, this limitation is significantly mitigated when using large batch sizes, where the probability of at least one positive pair per class appearing in each batch increases. 
Therefore, we recommend the use of sufficiently large mini-batches when applying SupCon in this context.

\section{Experiments and Results}
We apply the previously introduced adaptation of the SupCon loss to the Imitation Learning (IL) setting as an auxiliary objective that regularizes the supervised predictive losses and shapes the latent space at the output of the feature extractor layers. The baseline model \cite{amadori2024robust} is an end-to-end policy network based on Energy-Based Models (EBMs). As a primary testbeds, we train an agent to play the Sky Garden level of Astro Bot\footnote{https://www.teamasobi.com/games/astro-bot} and the Phrike boss fight of Returnal\footnote{https://housemarque.com/games/returnal}. The two titles offer different challenges for the agents. Astro Bot is a third-person 3D platformer that requires precise control of the character for platforming and vertical structures exploration. Returnal is a third-person 3D shooter that requires both advanced camera control for aiming and timely character control to avoid incoming attacks. Training datasets consist of approximately 3 hours of gameplay demonstrations per title.

In addition to Astro Bot and Returnal, we evaluate our approach in three Atari games: Ms. Pac-Man, Montezuma’s Revenge, and Space Invaders. These are problems more directly aligned with the assumptions of SupCon, due to their purely discrete action spaces. Moreover, they have simpler dynamics compared to Astro Bot and Returnal: they are 2D games with less input complexity and precision required, with fewer number of challenges.
The baseline model is composed of a feature extractor followed by a classifier layer trained with cross-entropy loss.
The code of this baseline for the Atari experiments is available \cite{kanervisto2020benchmarking}, along with the corresponding datasets from Atari-HEAD \cite{zhang2020atari}.

\subsection{Atari experiments}
For the Atari experiments, we trained five models using the best-performing set of hyperparameters and evaluated each resulting policy over 500 rollouts. 
We report average performance across these policies as a percentage improvement over the baseline, using the score of a random agent as the lower bound for normalization.

\begin{table}[]
\caption{Percentage of score improved training with SCIL with respect to the baseline}
\label{tab:atari}
\resizebox{\columnwidth}{!}{%
\begin{tabular}{c|cc|cc|cc}
Environment & \multicolumn{2}{c|}{Space Invaders} & \multicolumn{2}{c|}{Montezuma's   Revenge} & \multicolumn{2}{c}{Ms. Pac-Man} \\ \hline
 & Avg & Std & Avg & Std & Avg & Std \\
Increased score (\%) & 3.33 & 5.00 & 26.53 & 14.80 & 33.47 & 30.82
\end{tabular}%
}
\vspace{\vspacereducer}
\end{table}

\begin{figure}[htbp]
  \centering
  \begin{subfigure}{0.48\columnwidth}
    \centering
    \includegraphics[width=\linewidth]{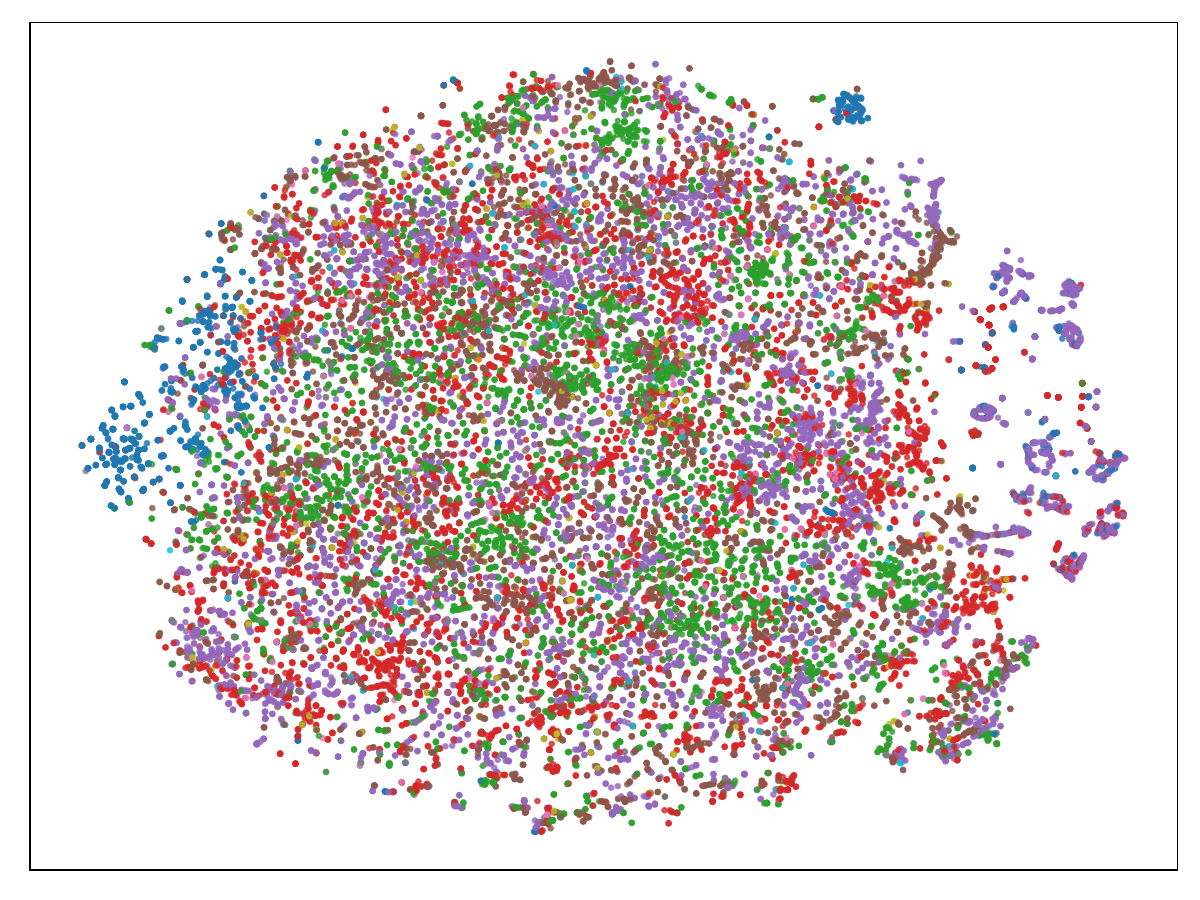}
    \caption{Baseline}
    \label{fig:fig1a}
  \end{subfigure}
  \hfill
  \begin{subfigure}{0.48\columnwidth}
    \centering
    \includegraphics[width=\linewidth]{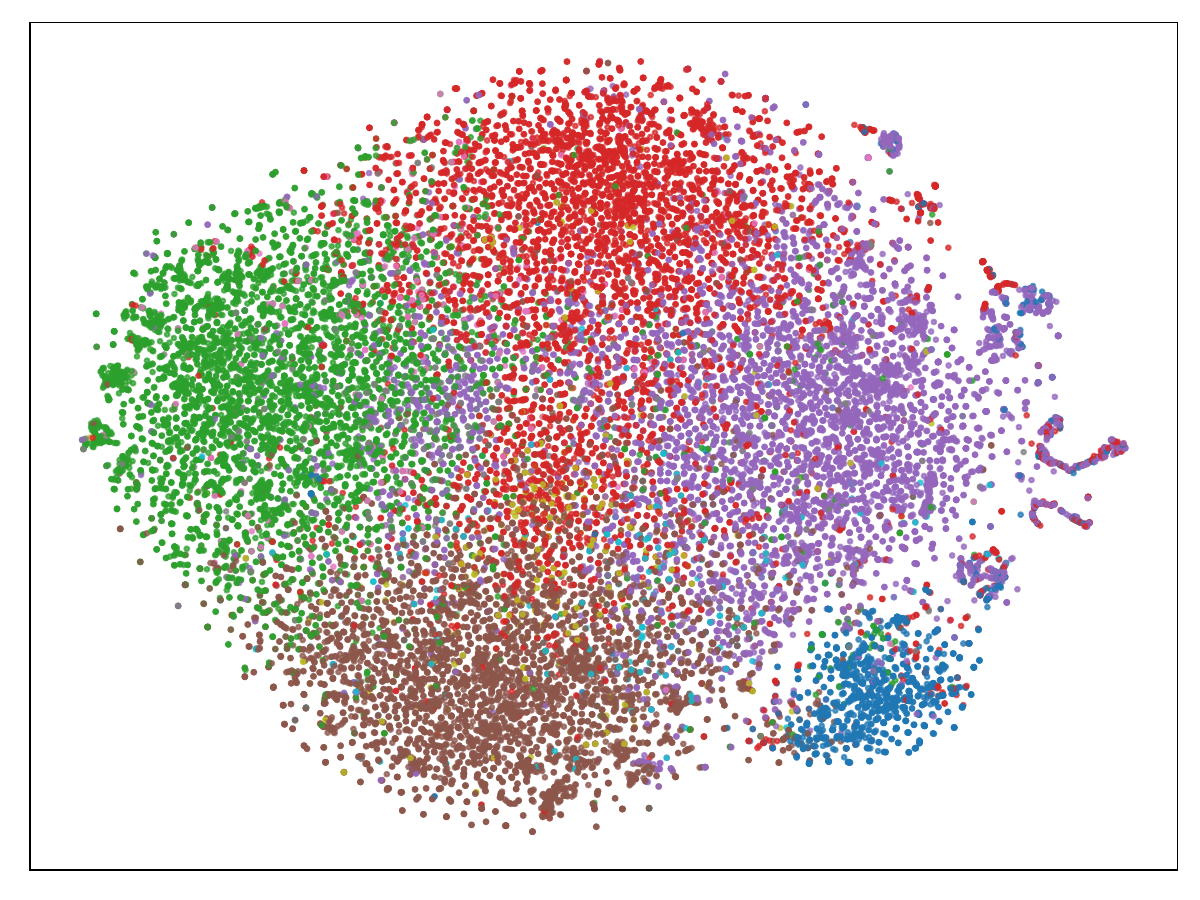}
    \caption{With SCIL (SupCon)}
    \label{fig:fig1b}
  \end{subfigure}
  \caption{Learned embedding spaces for Ms. Pac-Man.}
  \label{fig:embeddingsatari}
\vspace{\vspacereducer}
\end{figure}

Table \ref{tab:atari} shows that the percentage of policy improvement is positive in the tested games when the embeddings space is structured with the information of the output labels.
A comparison of such structured resulting embedding space is plotted in Fig. \ref{fig:embeddingsatari} for the Ms. Pac-Man environment.

\subsection{3D games experiments}
For Astro Bot, the level used for the evaluation includes a sequence of checkpoints that the agent must reach to complete the level successfully. For a more fine-grained assessment, we evaluate policy performance across individual segments of the level, where each segment begins at checkpoint $k$ and ends at checkpoint $k+1$, thus, the performance metric used is the success rate.
For Returnal, we measure the policy performance based on the percentage of damage done to the boss in a boss fight, wherein 100\% of damage means finishing a phase.

We trained three models using the proposed SCIL and three using the baseline. Each model was evaluated by running 10 rollouts per segment and measured the policy performance metric in order to compare both approaches.



\begin{figure}[htbp]
  \centering
  \begin{subfigure}{0.48\columnwidth}
    \centering
    \includegraphics[width=\linewidth]{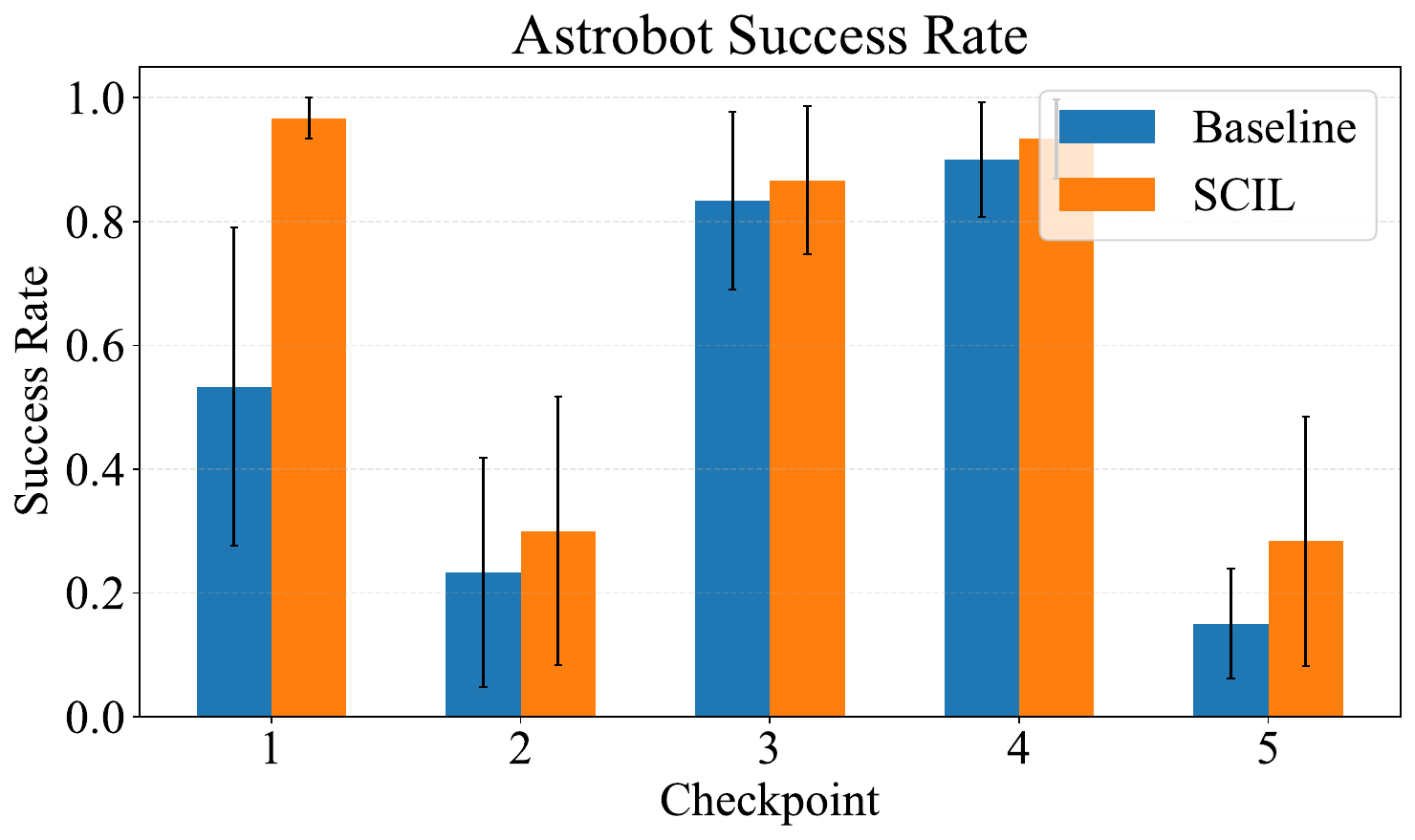}
    \label{fig:fig1a}
  \end{subfigure}
  \hfill
  \begin{subfigure}{0.48\columnwidth}
    \centering
    \includegraphics[width=\linewidth]{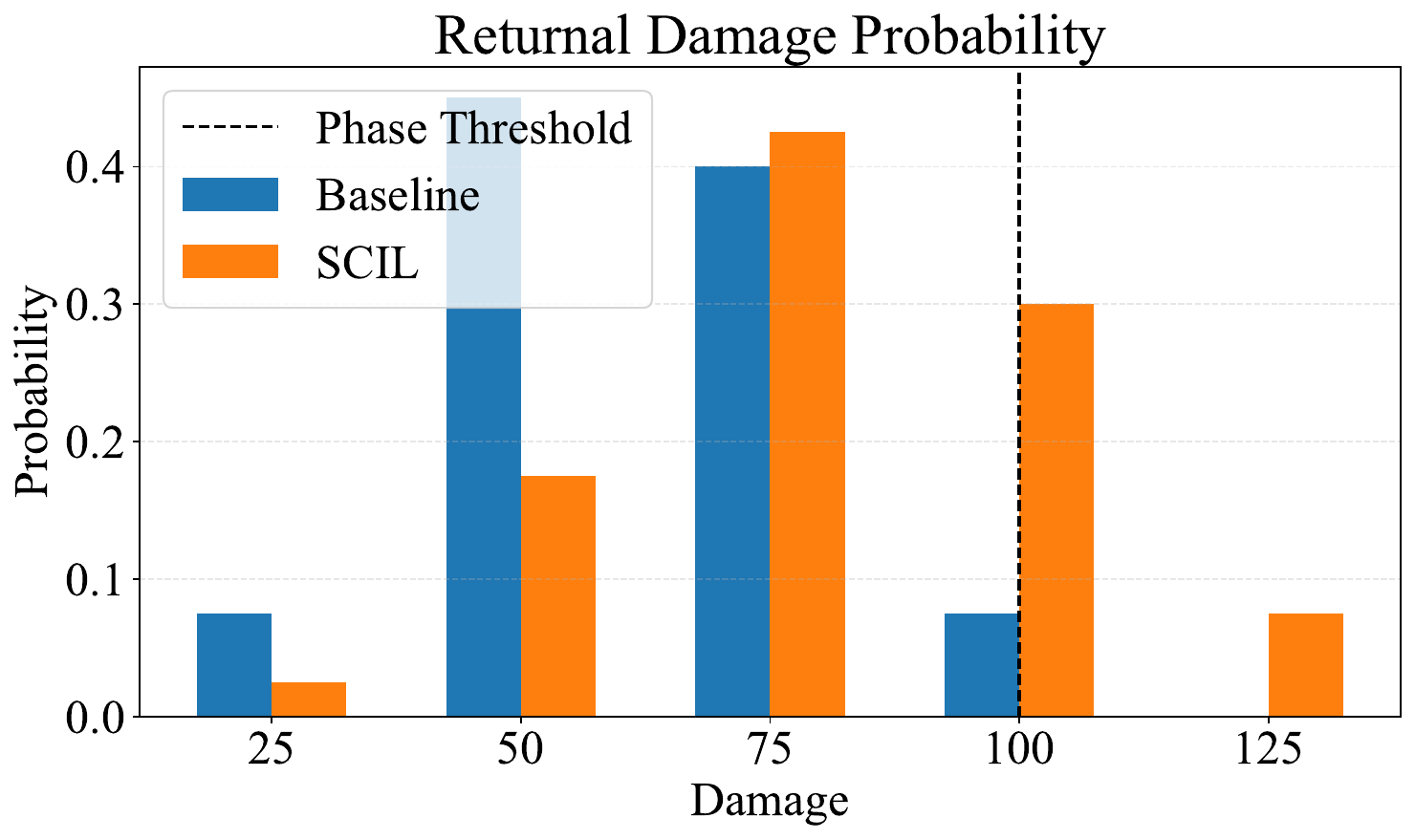}
    \label{fig:fig1b}
  \end{subfigure}
\vspace{\vspacereducer}
  \caption{Astro Bot success rate and Returnal damage probability, BL versus SCIL. The vertical bar in the Returnal chart indicates the damage required to transition to the second phase of the fight.}
  \label{fig:results_3Dgames}
\vspace{\vspacereducer}
\end{figure}

\begin{figure}[htbp]
  \centering
  \begin{subfigure}{0.48\columnwidth}
    \centering
    \includegraphics[width=\linewidth]{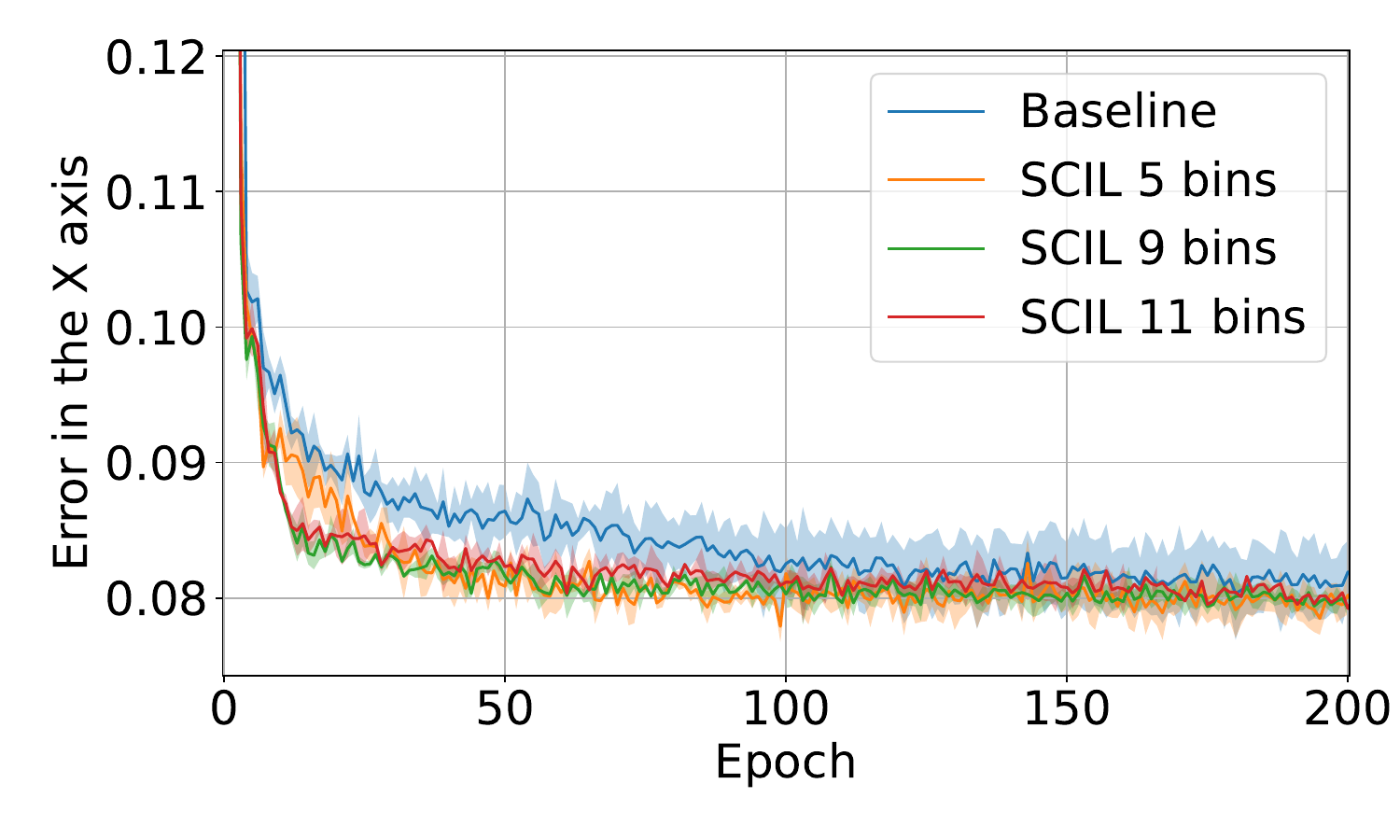}
    \label{fig:fig1a}
  \end{subfigure}
  \hfill
  \begin{subfigure}{0.48\columnwidth}
    \centering
    \includegraphics[width=\linewidth]{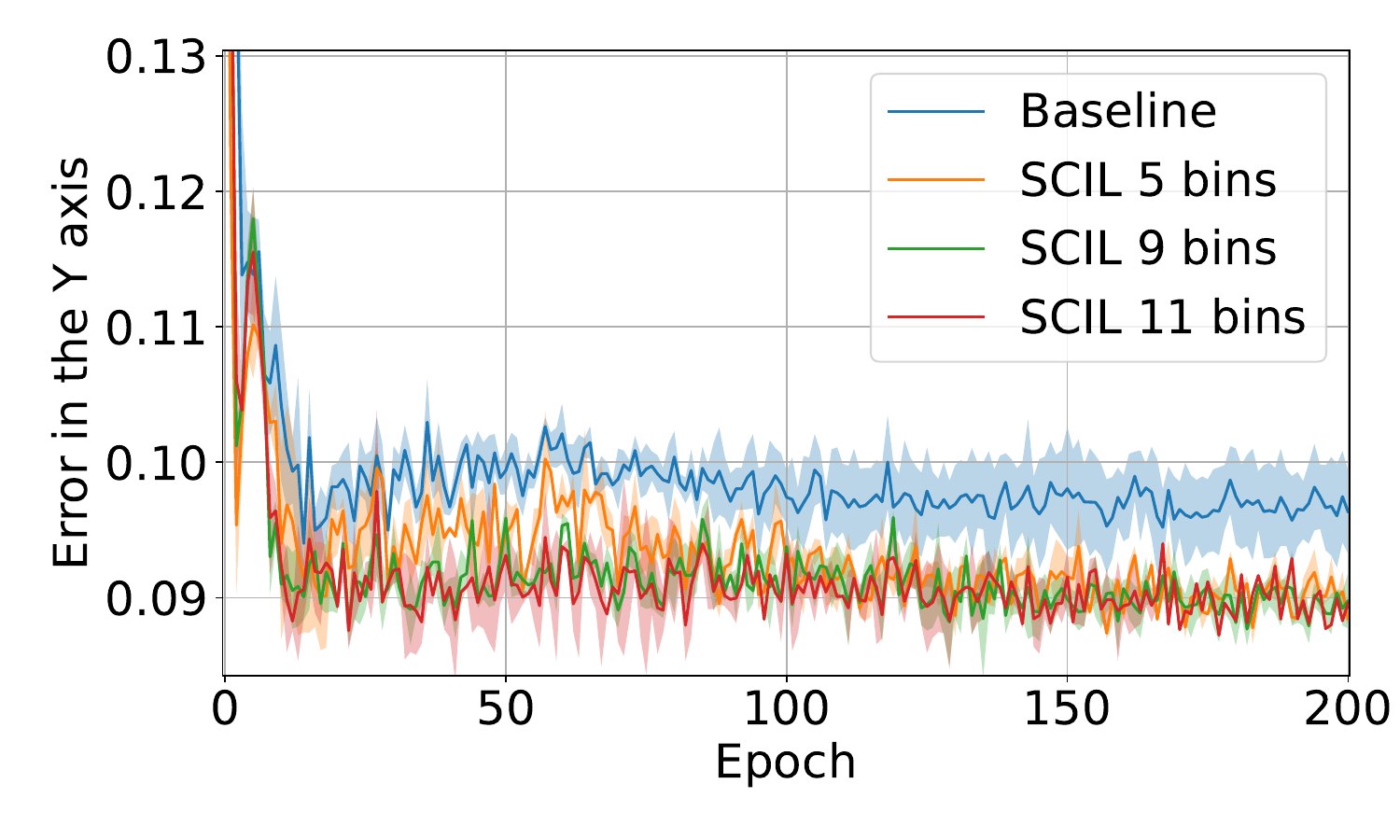}
    \label{fig:fig1b}
  \end{subfigure}
\vspace{\vspacereducer}
  \caption{Validation error in the joystick axes used for navigation. BL stand for baseline.}
  \label{fig:continuousvalidation}
\vspace{\vspacereducer}
\end{figure}

Figure \ref{fig:results_3Dgames} shows that for Astro Bot although some checkpoints are more challenging than others, the policies regularized with SupCon succeed more often than the baseline.
For Returnal, the same plot shows that the policies trained with SCIL tend to finish the fights with higher damage done to the boss (the more right skewed the distribution, the better). 
The policies trained with SCIL were able to finish the first phase of the boss fight in 37.5\% of the trials, whereas the baseline models succeeded only 7.5\% of the times.
While policy performance remains the primary evaluation metric for this learning setup, tracking the validation error over training epochs also provides insight into the generalization behavior of the proposed approach. 
As shown in Figure \ref{fig:continuousvalidation}, we observe improved generalization across different levels of action discretization used for the SupCon loss, where the number of bins $B$ determines the granularity of the continuous action encoding. 
These plots also show a faster convergence when using Supcon.
Video footage of the agents.\footnote{https://www.youtube.com/watch?v=z-c-1HamiJU}




\section{Conclusions}

The experimental results demonstrate consistent improvements in policy learning when applying SCIL. 
The proposed approach was evaluated across both 2D Atari games and the 2 complex 3D game environments, the latter presenting greater challenges in terms of high-dimensional observations, multi-modal and multi-dimensional action spaces (including continuous and discrete actions), and the need to learn a diverse set of skills.

The study also validated the effectiveness of the method across two different model architectures: one based on traditional classification outputs, and another based on Energy-Based Models (EBMs). Even in the worst-performing runs, the proposed method consistently outperformed the baselines in terms of policy performance metrics.

These findings support the central hypothesis: that observations corresponding to similar actions should be represented by similar embeddings (results of Fig. \ref{fig:embeddingsatari} match the description of the ideal embedded representations in Fig. \ref{fig:embeddings}). 
Enforcing this structure in the latent space leads to improved generalization and more data-efficient learning in Imitation Learning agents.

We believe that the insights from this work extend beyond video game applications and may prove valuable for IL agents in other domains, such as robotics.

\bibliographystyle{IEEEtran}
\bibliography{biblio}

\appendices
\section{Loss Function Code}
The SupCon loss implemented in this paper is based on the original implementation \cite{khosla2020supervised} with the additional changes introduced in Section \ref{sec:method}.

\begin{lstlisting}[language=Python
%, caption=SupCon loss
]
import torch
from torch import nn as nn
from torch.nn import functional as F


class SupConLoss(nn.Module):
    """
    SupCon loss definition:
    https://arxiv.org/pdf/2004.11362.pdf
    Code adapted from the one of Yonglong Tian (yonglong@mit.edu)
    """
    def __init__(self, device, temperature=0.07, base_temperature=0.07, n_bins=5):
        super(SupConLoss, self).__init__()
        self.temperature = temperature
        self.base_temperature = base_temperature
        self.n_bins = n_bins
        self.device = device

    def discretized_labels(self, actions):
        # how many bins to discretize per action
        granularity = torch.tensor(actions.shape[1]*[self.n_bins]).to(actions.device)
        #discretize each action
        discretized = (actions * (granularity - 1)).round().long().squeeze(dim=1)
        #compute positional encoding
        multipliers = torch.cat((torch.tensor([1]).to(actions.device),granularity[:-1].cumprod(0)))
        labels = (discretized * multipliers.unsqueeze(0)).sum(dim=1)
        return labels


    def forward(self, features, raw_actions):
        if len(features.shape) < 2:
            raise ValueError('`features` needs to be [bsz, ...],'
                             'at least 2 dimensions are required')

        batch_size = features.shape[0]
        labels = self.discretized_labels(raw_actions)
        labels = labels.contiguous().view(-1, 1)

        if labels.shape[0] != batch_size:
            raise ValueError('Num of labels does not match num of features')
        mask = torch.eq(labels, labels.T).float().to(self.device)

        # normalize features
        features = F.normalize(features, p=2, dim=1)
        # compute logits
        anchor_dot_contrast = torch.div(
            torch.matmul(features, features.T),
            self.temperature)
        # for numerical stability
        logits_max, _ = torch.max(anchor_dot_contrast, dim=1, keepdim=True)
        logits = anchor_dot_contrast - logits_max.detach()

        # mask-out self-contrast cases
        logits_mask = torch.scatter(
            torch.ones_like(mask),
            1,
            torch.arange(batch_size ).view(-1, 1).to(self.device),
            0
        )
        mask = mask * logits_mask

        # compute log_prob
        exp_logits = torch.exp(logits) * logits_mask
        eps = 1e-12
        log_prob = logits - torch.log(exp_logits.sum(1, keepdim=True) + eps)

        # compute mean of log-likelihood over positive
        # modified to handle edge cases when there is no positive pair
        # for an anchor point.
        mask_pos_pairs = mask.sum(1)
        mask_pos_pairs = torch.where(mask_pos_pairs < 1e-6, 1, mask_pos_pairs)
        mean_log_prob_pos = (mask * log_prob).sum(1) / mask_pos_pairs

        # loss
        loss = - (self.temperature / self.base_temperature) * mean_log_prob_pos
        loss = loss.view(1, batch_size).mean()

        return loss
\end{lstlisting}

\end{document}